\pdfoutput=1

\documentclass[11pt]{article}

\usepackage[]{acl}

\usepackage{times}
\usepackage{latexsym}
\usepackage{booktabs}
\usepackage{arydshln}
\usepackage{graphicx}
\usepackage{amsmath}
\usepackage{appendix}
\usepackage{multirow}
\usepackage{bm}
\usepackage[T1]{fontenc}
\usepackage{xcolor}

\usepackage[utf8]{inputenc}

\usepackage{microtype}

\title{Improving Bot Response Contradiction Detection via Utterance Rewriting}

\author{
Di Jin, Sijia Liu, Yang Liu, Dilek Hakkani-Tur\\
Amazon Alexa AI\\
\texttt{\{djinamzn,sijial,yangliud,hakkanit\}@amazon.com}
}

\begin{document}
\maketitle
\begin{abstract}
Though chatbots based on large neural models can often produce fluent responses in open domain conversations, one salient error type is contradiction or inconsistency with the preceding conversation turns. Previous work has treated contradiction detection in bot responses as a task similar to natural language inference, e.g., detect the contradiction between a pair of bot utterances. However, utterances in conversations may contain co-references or ellipsis, and using these utterances as is may not always be sufficient for identifying contradictions. This work aims to improve the contradiction detection via rewriting all bot utterances to restore antecedents and ellipsis. We curated a new dataset for utterance rewriting and built a rewriting model on it. We empirically demonstrate that this model can produce satisfactory rewrites to make bot utterances more complete. Furthermore, using rewritten utterances improves contradiction detection performance significantly, e.g., the AUPR and  joint accuracy scores (detecting contradiction along with evidence) increase by 6.5\% and 4.5\% (absolute increase), respectively.
\end{abstract}

\section{Introduction}

Latest chatbots powered by large pre-trained neural models have shown decent capabilities to maintain fluent and interesting conversations with human users~\citep{paranjape2020neural,roller-etal-2021-recipes,bao-etal-2021-plato,konrad2021alquist}. 
However, they are still prone to  various kinds of annoying mistakes~\citep{xu2020recipes,see-manning-2021-understanding,xu-etal-2021-bot}. 
One such error is contradiction or inconsistency, 
as illustrated in Table~\ref{tab:examples}. 

In order to reduce contradiction errors, one approach is to develop a detection model to identify such problems after a system produces response candidates. 
To this end, ~\citet{welleck-etal-2019-dialogue} characterized the modeling of persona-related consistency as a natural language inference (NLI) problem and constructed a dialog NLI dataset based on Persona-Chat. To cover a broader range of consistency types (e.g., persona, logic, causality, etc), \citet{nie-etal-2021-like} collected DECODE, a dataset containing human written dialogues with self-contradictory utterances.
Besides the in-distribution  human-human dialogues test set, they collected an out-of-distribution set containing dialogues between human and different chatbots. 
This human-bot test set can better evaluate models' performance in detecting contradiction in conversations between human and chatbots, which is the focus of this work.

\begin{table*}[t]
\small
\centering
\resizebox{0.9\textwidth}{!}{
\begin{tabular}{p{1.1cm}|p{6.3cm}|p{6.4cm}|p{2.3cm}}
\toprule
\textbf{Speakers} & \textbf{Original Dialogues} & \textbf{Rewritten Dialogues} & \textbf{Rewriting Type} \\ \midrule
Human: & Hi, what's your favorite singer? & --- \\
Bot: & \textbf{Mine} is johnny cash of course. & \textbf{My favorite singer} is Johnny Cash, of course. & Co-reference \\
Human: & He's amazing, I love his songs. & --- \\
Bot: & \noindent{\color{red} I never got to see johnny cash play but I wish I did.} & I never got to see Johnny Cash play but I wish I did. & None \\
Human: & Same, I wish I could go to one of his concerts. & --- \\
Bot: & \noindent{\color{red}I have not been since last year though. I like sports.} & I have not been \textbf{to Johnny Cash's concert} since last year though. I like sports. & Ellipsis \\ \bottomrule
\end{tabular}}
\caption{Examples of human-bot conversations with contradictory bot utterances marked by red color. We rewrite every bot utterance to restore co-references and ellipsis (the restored parts are highlighted by bold font). }
\label{tab:examples}
\vspace{-3mm}
\end{table*}

We find one failure of the state-of-the-art (SOTA) contradiction detection model is due to the frequent anaphora and ellipses in chatbot utterances.
One typical example is shown in Table~\ref{tab:examples}, where the first bot utterance has an anaphor, ``mine'', and the last bot utterance misses an important entity, ``Johnny Cash's concert''. 
Such incomplete utterances would prevent detection models from fully understanding the bot utterances in the dialog, thus leading to detection errors. Therefore, we propose to first rewrite the bot utterances to recover all the missing information and then perform the contradiction detection task. 
To support this goal, we first collect a new dataset for incomplete utterance rewriting, which is a widely studied task~\citep{pan-etal-2019-improving,su-etal-2019-improving,hao-etal-2021-rast} but still lacks supporting datasets for open-domain conversations in English~\citep{quan-etal-2019-gecor}. Then we propose a rewriting model trained on this data to rewrite the anaphors to their corresponding entities and restore any missing content. 
We conduct experiments on the DECODE dataset~\citep{nie-etal-2021-like}, and demonstrate substantial performance improvement in contradiction detection when the utterance rewrite module is applied. 
Overall, we have made the following contributions in this work:

\begin{itemize}
  \vspace{-0.05in}
   \item We have collected a new English dataset for incomplete utterance rewriting for general open-domain conversations, and developed a rewriting model for utterance restoration. 
   \vspace{-0.05in}
    \item With bot utterance rewriting, we can improve the previous best contradiction detection model by 6.5\% in AUPR and 4.5\% in joint accuracy that considers both contradiction and evidence labels.
  \vspace{-0.05in}    
    \item We relabeled the  human-bot test set of the benchmark DECODE dataset and corrected some annotations.\footnote{Code and data are released at: https://github.com/jind11/utterance-rewriting}
    
\end{itemize}

\section{Contradiction Detection Method}

\subsection{Task Definition}

We formalize dialogue contradiction detection as an NLI task. Given a list of utterances $x=\{u_1^H,u_1^B,...,u_n^H,u_n^B\}$ representing a dialogue, the task is to determine if the last bot utterance $u_n^B$ contradicts any previously conveyed information contained in the past bot utterances 
$\{u_1^B,...,u_{n-1}^B\}$. 
Note that we are using human and bot alternating turns here (referred to as H and B), but they can be human-human conversations too. 
In addition to the binary label $y$, with 0 or 1 corresponding to the non-contradiction and the contradiction labels, respectively, 
we also output a set of indices $I\in \{1,...,n-1\}$ representing the utterances in $\{u_1^B,...,u_{n-1}^B\}$ that is actually contradicted by the last utterance $u_n^B$. 

\subsection{Detection Models}
\label{sec:detection-models}

Based on the benchmark DECODE dataset, \citet{nie-etal-2021-like} proposed two approaches for contradiction detection: an unstructured approach and a structured utterance-based (SUB) approach. The former one concatenates all the previous utterances in the dialogue history to form a single textual context. Then a classification model $f_\theta$ is applied to the context and the last utterance to infer the probability of contradiction.
The latter \textsc{SUB} approach pairs every past bot utterance with the last one, and then feeds each pair to the classification model $f_\theta^{SUB}$. The final contradiction probability is the maximum over all the outputs: $\hat{y}=max\{f_\theta^{SUB}(u_i^B,u_n^B): i\in \{1,...,n-1\}\}$. 
The supporting evidence (SE) for a contradiction decision contains the pairs having contradiction probability higher than a threshold $\eta$, i.e., $I=\{i:f_\theta^{SUB}(u_i^B,u_n^B)>\eta\}$. 
\citet{nie-etal-2021-like} demonstrated that the latter \textsc{SUB} approach significantly outperforms the former one on the human-bot test set (more than 10\% in accuracy). This \textsc{SUB} method is \textbf{the current SOTA model for contradiction detection}, which we adopted as one baseline.

\vspace{-0.05in}
\subsection{Utterance Rewriting for Contradiction Detection}

As discussed earlier, we noticed that many bot utterances contain co-references and ellipses and thus the baseline model fails to capture the semantic meaning or contradiction in the sentence pair. 
Therefore, 
we propose to first rewrite the bot utterances to restore co-references and ellipsis, and then feed the rewritten utterances (e.g., the dialogues on the right in Table \ref{tab:examples}) to the model. 
To this end, we first collect a new dataset specially for utterance rewriting and then develop a rewriting model.

\paragraph{Rewriting Data Collection}

To get parallel training data for utterance rewriting for open-domain conversations, we sub-sampled 6,000 and 4,000 dialogues from the DailyDialog~\citep{li-etal-2017-dailydialog} and BST~\citep{smith-etal-2020-put} datasets, respectively, as the training set. Besides, we sub-sampled 400 and 400 dialogues from DailyDialog and BST, respectively, as the test set. We only use the first six utterances in each dialog. Specifically, we use the first two utterances (from both speakers) as leading context and ask annotators to check the remaining four utterances, following \citet{pan-etal-2019-improving}.\footnote{Utterance rewriting needs context to resolve co-references and ellipsis, and thus the first two utterances are not suitable for rewriting annotation.} Empirically we find that the context information needed to resolve co-references and ellipsis can always be found within 1-3 turns~\citep{pan-etal-2019-improving,su-etal-2019-improving}. We ask annotators to identify whether an utterance is complete and can be understood without reading the context, and if not, then rewrite it to restore any missing information. 

To ensure the annotation quality, we hired three in-house professional data annotators, who have been first trained via a pilot annotation session and then proceed to the official annotation phase after passing our provided qualification set. In the official annotation phase, two of them first worked independently and then the third annotator was tasked to  make the adjudication over the two annotations and pick the best one or make revisions if needed. Besides, we  periodically sampled 10\% of the annotations from each annotator throughout the annotation process and provided feedback. The annotation is considered valid only when the accuracy of examined results surpassed 95\% (we deem those rewrites that are both correct and complete as correct rewrites, and then calculate the percentage of correct rewrites as the accuracy).
Overall, we have obtained 40,000 and 3,200 samples for training and testing, respectively. 

\paragraph{Rewriting Model}
We treat rewriting as a sequence-to-sequence (Seq2Seq) task and adopt two pre-trained Seq2Seq models, T5~\citep{raffel2019exploring} and Pegasus~\citep{zhang2019pegasus}. The input is the concatenated  context utterances and the original last utterance, with special tokens  inserted before each utterance to indicate its speaker.  

\vspace{-0.05in}
\section{Experiments}

\subsection{Contradiction Detection Data}

We use the DECODE dataset~\citep{nie-etal-2021-like} in this study.
However, we found some issues with its human-bot test set: (1) Around one third of non-contradiction dialogues contain only one human and one bot utterances, which makes the detection task over-simplified, since there are no previous bot utterances. (2) Not every bot utterance has been annotated for contradiction with respect to its history. 
(3) Evidence is not labeled to indicate which history bot utterance contradicts the last one. 

To resolve the above-mentioned issues, we  curate new annotation using the dialogues in the original test set.
Details of annotation procedures are provided in Section \ref{sec:appendix-data-collection} of the Appendix. Overall, we have obtained 1,889 samples (453 positive samples and 1,436 negative ones), which we call an unbalanced set. Besides, we sub-sampled 453 negative samples and combined them with all the positive ones to form a balanced set. Table~\ref{tab:data-stats} summarizes the data statistics.
We will release this new test set. 

\subsection{Baselines}

We compare the contradiction detection performance with and without rewriting bot utterances, all based on the same \textsc{SUB} model framework, which is the current SOTA model for contradiction detection. Another baseline we introduced is \textsc{SUB-Concat}, where each bot utterance is the concatenation of the original one with the preceding human utterance such that the missing information (coreference or ellipsis) can be recovered from the included previous utterance.  

For rewriting, we compare our model against four strong baselines: one is the off-the-shelf SOTA co-reference resolution model trained on OntoNotes (named as ``Co-reference'')~\cite{toshniwal-etal-2021-generalization,wu-etal-2020-corefqa}, and the other three are developed based on three related datasets for rewriting, named as ``CANARD''~\cite{elgohary-etal-2019-unpack}, ``Gunrock''~\cite{zhang2020filling}, and ``MuDoCo''~\cite{tseng-etal-2021-cread}. Specifically, CANARD is a query rewriting dataset that aims to rewrite a query/question based on previous consecutive QA pairs for the conversational question answering task. The Gunrock dataset focuses on resolving ellipsis while containing a small portion of co-reference cases, and it consists of 1745 samples where all dialogues are in-house curated following the Alexa Prize competition format. The MuDoCo dataset is also for query rewriting for task-oriented dialogues covering 6 domains.


\subsection{Evaluation Metrics}
\label{sec:evaluation-metrics}

To evaluate incomplete utterance rewriting, we use both automatic and human evaluation. For human evaluation, we propose two metrics: (1) Correctness; (2) Completeness. The former one checks whether the rewriting part is correct and obeys the information in dialogue context, while the latter one checks whether the rewritten utterance is complete enough to be understood without reading the context. We have binary labels for both metrics and report the percentage of positive labels after human evaluation. For automatic evaluation, in addition to the widely used  BLEU~\citep{papineni2002bleu}, ROUGE-1 (R-1), and ROUGE-L (R-L)~\citep{lin-2004-rouge}, we have added two more metrics specially for evaluating text editing models: exact match (\textsc{EM}) accuracy, and the $F_1$ score, which was proposed in \citet{pan-etal-2019-improving} and focuses on n-grams that contain at least one restored word. Specifically, the n-gram restoration precision, recall, and F-score can be calculated as:

\vspace{-0.2in}
\begin{equation*}
\centering
\begin{aligned}
    P_n=\frac{\{\textrm{restored n-grams}\}\cap \{\textrm{n-grams in ref}\}}{\{\textrm{restored n-grams}\}} \\
    R_n=\frac{\{\textrm{restored n-grams}\}\cap \{\textrm{n-grams in ref}\}}{\{\textrm{n-grams in ref}\}} \\
    F_n=2\cdot\frac{P_n*R_n}{P_n+R_n}
\end{aligned}
\vspace{-0.1in}
\end{equation*}
where ``restored n-grams'' refer to the n-grams in restored utterance that contain at least one restored words, and ``n-grams in ref'' refer to the n-grams in reference that contain at least one restored words.

For contradiction detection, we first set the threshold $\eta$ to be 0.5, and report Precision/Recall/F1 for both the binary contradiction label and the support evidence labels, following \citet{thorne-etal-2018-fact}.\footnote{https://github.com/sheffieldnlp/fever-scorer} Besides, we report Joint Accuracy, which indicates the performance when both the 2-way contradiction detection and the supporting evidence retrieval are correct. Considering that these scores are sensitive to $\eta$, we also report Area-under-Precision-Recall-Curve (AUPR) as a threshold-independent score. 

\subsection{Experimental Setup}


For utterance rewriting, we have used three kinds of pre-trained models: T5-Base, T5-Large, and Pegasus-Large, whose parameter sizes are 220 M, 770 M, and 568 M. Each model is trained for 4 epochs with a learning rate of $5e^{-5}$, and beam search (beam size of 5) is used for generation. 

For contradiction detection, following \citet{nie-etal-2021-like}, we used the RoBERTa-Large model whose parameter number is 330 M, which is trained for 3 epochs with a learning rate of $1e^{-5}$. We have used the Huggingface Transformer code base\footnote{https://github.com/huggingface/transformers/tree/master} and all experiments were run on Nvidia V100 GPUs.

\vspace{-0.05in}
\section{Results and Discussion}
\vspace{-0.05in}
\subsection{Utterance Rewriting}

We performed both automatic and human evaluation for utterance rewriting (please refer to  Section \ref{sec:evaluation-metrics} for evaluation details). Table~\ref{tab:rewriting-automatic-eval} summarizes the automatic evaluation results. As can be seen, the three models perform similarly overall, with T5-Large slightly outperforming the other two. We thus adopt it as the main rewriting model in later experiments.

\begin{table*}[t]
\small
\centering
\resizebox{0.99\textwidth}{!}{
\begin{tabular}{ll|cccc|ccc}
\toprule
\multirow{2}{*}{\textbf{Detection Method}} & \multirow{2}{*}{\textbf{Rewriting Model}}   & \multicolumn{4}{c|}{\textbf{Balanced Set}}                                                          & \multicolumn{3}{c}{\textbf{Unbalanced Set}}                                 \\
                           & & \textbf{P/R/F1}                   & \textbf{AUPR}                 & \textbf{SE (P/R/F1)}          & \textbf{Joint-Acc.}           & \textbf{P/R/F1}                   & \textbf{AUPR}                 & \textbf{Joint-Acc.}           \\ \midrule
\textsc{SUB}-Bot only          & \multirow{2}{*}{None} & 89.4/70.6/78.9                & 89.0                & 90.4/62.9/74.2   & 69.9                & 73.2/70.6/71.9                & 75.4                & 81.4                \\
\textsc{SUB}-Concat            & &    88.1/66.9/76.0                  &  88.1 &    90.0/60.0/72.0 & 68.1 & 66.3/66.9/66.6 & 71.6 &    78.7                  \\ \midrule
\multirow{4}{*}{\textsc{SUB}-Bot only} & Co-reference & 89.0/71.3/79.2 & 89.1 & 90.0/64.5/75.2 & 69.7 & 73.1/71.3/72.2 & 75.8 & 81.3 \\
 & CANARD & 79.3/37.1/50.5 & 73.9 & 89.6/26.3/40.7 & 54.6 & 60.0/37.8/46.3 & 52.8 & 74.8 \\
 & Gunrock & 79.2/59.8/68.2 & 74.0 & 88.5/50.2/64.0  & 60.0 & 53.0/59.8/56.2 & 49.0 & 71.9 \\
 & MuDoCo & 88.1/65.3/75.0 & 87.4 & 91.6/59.5/72.1 & 67.9 & 70.6/65.3/67.9 & 71.2 & 80.2 \\ \midrule
\multirow{2}{*}{\textsc{SUB}-Bot only} & Ours (single) & \underline{90.9}/\textbf{72.9}/\underline{80.9}                & \underline{91.8}                & \underline{93.0}/\textbf{\underline{67.6}}/\textbf{\underline{78.3}}    & \underline{73.6}                & \underline{73.5}/\textbf{72.9}/\underline{73.2}                & \underline{78.6}                & \underline{82.8}                \\
     & Ours (ensemble) & \textbf{92.9}/{71.7}/\textbf{81.0} & \textbf{93.2} & \textbf{93.9}/66.1/77.6 & \textbf{74.4} & \textbf{80.1}/71.7/\textbf{75.7} & \textbf{81.9} & \textbf{84.8}   \\ \bottomrule
\end{tabular}}
\vspace{-2mm}
\caption{Contradiction detection performance (\%) on new human-bot test set. Best results for single model (T5-Large) are marked by underlines while overall best results are marked bold. \textsc`{SUB}-Bot only' means feeding only bot utterances to the \textsc{SUB} model while \textsc{SUB}-Concat uses the concatenated bot utterance and the preceding human turn. 
}
\vspace{-4mm}
\label{tab:contradiction-performance}
\end{table*}

\begin{table}[!htpb]
\centering
\small
\resizebox{0.99\columnwidth}{!}{
\begin{tabular}{lccccc}
\toprule
\textbf{Models}        & \textbf{BLEU}                      & \textbf{R-1}                         & \textbf{R-L}                        & \textbf{EM} & {$\bm{F_1}$} \\ \midrule
T5-Base       & 0.653 & 0.822 & 0.801 & 0.213                 & 0.402                 \\
T5-Large      & 0.653 & 0.820 & 0.798 & 0.199                 & 0.422                 \\
Pegasus-Large & 0.649                     & 0.822                     & 0.801                     & 0.212                 & 0.391                \\ \midrule
Agreement & 0.714 & 0.840 & 0.837 & 0.323 & 0.309
\\ \bottomrule
\end{tabular}}
\vspace{-2mm}
\caption{Automatic evaluation results for the rewriting model on the rewriting test set. Agreement is the inter-annotator agreement between two rewrites in test set.}
\label{tab:rewriting-automatic-eval}
\vspace{-2mm}
\end{table}

\begin{table}[!hptb]
\centering
\small
\resizebox{0.99\columnwidth}{!}{
\begin{tabular}{lccc}
\toprule
\textbf{Test Set}      & \textbf{Correctness} & \textbf{Completeness} & \textbf{Change Rate} \\ \midrule
Rewriting     & 92.0        & 85.0         & 59.0        \\
Contradiction & 98.0        & 93.1         & 62.4       \\ \midrule
\textbf{Test Set}      & \textbf{Co-reference } & \textbf{Ellipsis } & \textbf{Incomplete } \\ \midrule
Rewriting     & 39.0        & 42.0         & 68.0         \\
Contradiction & 42.6        & 27.7         & 58.4       \\ 
\bottomrule
\end{tabular}}
\caption{Upper block: human evaluation of rewriting for both the rewriting and contradiction detection test sets (\%); bottom block: percentage of utterances containing co-reference or ellipsis, or either (incomplete). 
}
\vspace{-2mm}
\label{tab:rewriting-human-eval}
\vspace{-4mm}
\end{table}




We also sub-sample 100 rewritten utterances by T5-Large for human evaluation. 
As shown in Table~\ref{tab:rewriting-human-eval}, the correctness and completeness scores for both test sets are above 85\%, validating the high-quality of the rewriting model. 
We also report the change rate in the table that defines the percentage of the rewritten utterances that are different from the original ones (only differences in punctuation and upper/lower-case are not considered).
The bottom block of Table \ref{tab:rewriting-human-eval} shows the percentage of utterances containing co-reference or ellipsis, or either, i.e, incomplete utterances. We see that \textbf{co-reference and ellipsis occur almost equally frequently} in incomplete utterances.
Considering all the numbers together, we demonstrate that the rewriting model has covered most of those incomplete utterances. 

\vspace{-0.05in}
\subsection{Contradiction Detection}

Table \ref{tab:contradiction-performance} compares the contradiction detection performance without rewriting and with rewriting by different rewriting models. 
First of all, the \textsc{SUB}-Concat method without rewriting does not yield any performance gain although it has included the context utterances. More importantly, after rewriting all bot utterances for both training and test sets, only our rewriting model can lead to significant improvements for all the evaluation metrics, while those baseline rewriting models either maintain or deteriorate the performance (we provided the rewriting performance of these baselines in Section \ref{sec:appendix-other-rewriting-models} of Appendix for reference). We see that the AUPR metric has been improved by around 2.8\% and 3.2\% absolutely for the balanced and unbalanced sets by our model, respectively. 
We also implemented model ensemble where we rewrite bot utterances using our three rewriting models (T5-Base/Large and Pegasus-Large), run contradiction detection using each,  and average their contradiction scores to obtain the final prediction.
This further improves the detection performance over single models. 
Overall, we have achieved a substantial increase of 4.2\% and 6.5\% for AUPR and 4.5\% and 3.4\% for Joint-Acc. for the balanced and unbalanced sets, respectively. 

\vspace{-0.05in}
\subsection{Error Analysis}
\vspace{-0.05in}
We conducted additional error analysis to understand the performance gains and remaining errors.  
We first obtained 95 false negative samples by the ``\textsc{SUB}-Bot only'' model without applying rewriting, and then manually identified 28 samples whose last bot utterances are incomplete. 
We then manually rewrote those incomplete bot utterances. With such manual rewriting, we are able to correctly classify 18 out of 28 samples to be positive (64.3\% in accuracy), whereas, with the T5-Large rewriting model, 15 samples can be correctly predicted (53.6\% in accuracy). This comparison indicates that our automatic rewriting has pushed the performance improvement close to the upper bound achieved by manual rewriting. More error analysis is provided in Section \ref{sec:appendix-error-analysis} of Appendix.

\subsection{Why Utterance Rewriting Helps?}

As illustrated by Table 1, in order to infer the entailment relationship between the premise (i.e. ``Mine is johnny cash of course.'') and hypothesis (i.e. ``I have not been since last year though.''), we need to resolve the anaphora and ellipses so that some key information can be restored, e.g., ``Mine'' is replaced by ``My favorite singer'' in the premise and the missing phrase of ``to Johnny Cash's concert'' is restored in the hypothesis. Without restoring such key information from the dialogue context, the contradiction detection model cannot fully understand the premise and hypothesis sentences, thus not being able to accurately detect contradictory cases. One could argue that we can simply concatenate the context with both premise and hypothesis respectively so that the detection model could grab the missing information itself from the context, however, the baseline method ``SUB-Concat'', which follows this setting, still under-performs the baseline without concatenating the context (i.e. SUB-Bot only). This indicates that when the premise and hypothesis are organized in a dialogue structure with multiple turns rather than as single-turn sentences, the NLI based detection model is not good at inferring their relationship anymore. Therefore, we need to use the utterance rewriting model to grasp the most necessary information from context and insert into the bot utterances so that we can still use the single-turn format while making up the missing information for entailment inference. 

\subsection{Future Work}

We will keep improving the utterance rewriting model. Besides, we will showcase that utterance rewriting can also help improve other dialogue related tasks, such as task-orientated dialogue state tracking and response generation, open-domain dialogue response selection and generation, etc. 

\vspace{-0.05in}
\section{Conclusion}
\vspace{-0.07in}
In this work, we aim to improve contradiction detection in chatbot utterances via rewriting to restore anaphora and ellipsis. To develop such an utterance rewriting model, we curated a dataset by crowd-sourcing and demonstrated that the rewriting quality is satisfactory.
With such a rewriting technique, we are able to significantly improve the contradiction detection performance. 


\bibliography{custom}
\bibliographystyle{acl_natbib}

\newpage
\clearpage
\appendix

\setcounter{table}{0}
\setcounter{figure}{0}
\setcounter{footnote}{0}
\setcounter{page}{1}
\renewcommand\thetable{\Alph{section}.\arabic{table}}
\renewcommand\thefigure{\Alph{section}.\arabic{figure}}

\section{Contradiction Detection Data Collection}
\label{sec:appendix-data-collection}

Considering that the original human-bot test set of the benchmark DECODE dataset is problematic, we specially curate new annotation based on those dialogues of the original test set via the following steps: (1) We first obtained 507 unique and full dialogues from the original human-bot test set\footnote{https://github.com/facebookresearch/ParlAI/blob/main/\ projects/contradiction} by merging dialogues with overlaps and removing dialogues of only one turn. We then obtained 1,889 partial dialogues for annotation by cutting each full dialogue from the beginning to each bot utterance so that we can annotate whether each bot utterance contradicts against its context. (2) In the first round of annotation, we ask three Amazon Mechanical Turk workers (from English-speaking countries, including USA, England, and Canada) to annotate both the binary label of contradiction and evidence indices that indicate which history bot utterance contradicts the last one. When setting-up the annotation interface, we have provided one line of guidance to warn annotators not to reveal any personal information during annotation. We keep those samples with three full votes as finalized samples and pass those without three equal votes to the second round. (3) In the second round, we provide the maximum set of evidence indices to another three AMT workers and let them verify and write down new annotation if they do not agree. Again, samples with three agreements are selected as finalized ones and those without are passed to authors of this work for final adjudication. Finally, among all the 1,889 samples, we have obtained 453 positive samples and 1,436 negative ones, which we call an unbalanced set. Besides, we have also sub-sampled 453 negative samples and combine them with all positive ones to form a balanced set. Table~\ref{tab:data-stats} summarizes the data statistics.

\vspace{-0.07in}
\begin{table}[!htpb]
\centering
\small
\resizebox{0.99\columnwidth}{!}{
\begin{tabular}{lccc}
\toprule
Dataset         & Positive & Negative & Type        \\ \midrule
Train           & 13,592   & 13,592   & Human-Human \\
Balanced Test   & 453      & 453      & Human-Bot   \\
Unbalanced Test & 453      & 1,436    & Human-Bot  \\ \bottomrule
\end{tabular}}
\caption{Statistics of the contradiction detection dataset.}
\label{tab:data-stats}
\vspace{-4mm}
\end{table}

\section{Rewriting Quality of Baselines}
\label{sec:appendix-other-rewriting-models}

Table \ref{tab:rewriting-automatic-eval-baselines} compares our rewriting model with baselines that are developed on three related datasets for utterance rewriting (CANARD, Gunrock, and MuDoCo) and we report performance on our rewriting test set. As expected, our rewriting model that is trained on our own rewriting dataset performs the best. And by combining Table \ref{tab:contradiction-performance} and Table \ref{tab:rewriting-automatic-eval-baselines}, we can see that better rewriting performance on our rewriting test set can lead to better contradiction detection performance. 

\vspace{-0.07in}
\begin{table}[!htpb]
\centering
\small
\resizebox{0.99\columnwidth}{!}{
\begin{tabular}{lccccc}
\toprule
\textbf{Models}        & \textbf{BLEU}                      & \textbf{R-1}                         & \textbf{R-L}                        & \textbf{EM} & {$\bm{F_1}$} \\ \midrule
CANARD       & 0.363 & 0.631 & 0.623 & 0.105 & 0.215                 \\
Gunrock      & 0.352 & 0.626 & 0.612 & 0.008 & 0.223                   \\
MuDoCo & 0.576 & 0.783 & 0.781 & 0.214 & 0.133 \\ \midrule
Ours & 0.653 & 0.820 & 0.798 & 0.199                 & 0.422
\\ \bottomrule
\end{tabular}}
\caption{Automatic evaluation results for the rewriting model on the rewriting test set. Agreement is the inter-annotator agreement between two rewrites in test set.}
\label{tab:rewriting-automatic-eval-baselines}
\end{table}

\section{Qualitative Error Analysis}
\label{sec:appendix-error-analysis}

Among all 95 false negative samples predicted by the baseline, we find that the last bot utterances of 28 samples are incomplete and need rewriting for restoration. After automatic rewriting of all bot utterances, we can get 15 samples correct but still have 13 samples being false negatives. Now we would like to analyze the error pattern of these 13 false negatives after rewriting and we can categorize the errors into four types: numerical reasoning, logical reasoning, common sense reasoning, and hard to judge. Table \ref{tab:appendix-still-wrong-examples} provides examples for each type. Here are detailed definitions for these four types: (1) Numerical reasoning: Models need to do some numerical calculation or comparison to make a decision; (2) Logical reasoning: It requires logical reasoning for prediction; (3) Common sense reasoning: Some common sense knowledge needs to be equipped for reasoning; (4) Hard to judge: It is even hard for human to judge whether contradiction really exists or not.

Table \ref{tab:appendix-get-corrected-samples} provides several examples that are originally false negatives before rewriting bot utterances but later get correctly predicted after rewriting. As can be seen, the rewriting process can make up those critical information needed for detecting contradiction cases. Taking the first sample in Table \ref{tab:appendix-get-corrected-samples} as an example, there is ellipsis in the last two bot utterances in the original dialogue, which would lead to models' misunderstanding. After rewriting, the last two bot utterances get complete, which makes it much easier for models' decision making. 

\begin{table*}[t]
\centering
\small
\resizebox{0.99\textwidth}{!}{
\begin{tabular}{p{6.5cm}|p{6.5cm}|p{1.7cm}}
\toprule
\textbf{Original Dialogue}                                                                                                                                                                                                                                                                                                                                                                                                                                                                                                                                                                                      & \textbf{Rewritten Dialogue}                                                                                                                                                                                                                                                                                                                                                                                                                                                                                                                                                                                                    & \textbf{Error Type}           \\ \midrule
\textbf{human: }Hi!\newline \textbf{bot: }What do you do for a living? I'm retired, so I just sew all day.\newline \textbf{human: }I'm a homemaker. What kind of things do you sew?\newline \textbf{bot: }I sew a lot of quilts for my grandchildren. Do you have any kids?\newline \textbf{human: }Aww that's sweet! Yes, I have one son. \newline \textbf{bot: }How old is he? I have two daughters. They're a handful!\newline \textbf{human: }He's four and yes they are a handful! How old are your daughters?\newline \textbf{bot: }4, 5, and 8. I love them to death, but they're exhausting!                                                                             & \textbf{human: }Hi!\newline \textbf{bot: }What do you do for a living? I'm retired, so I just sew all day.\newline \textbf{human: }I'm a homemaker. What kind of things do you sew?\newline \textbf{bot: }I sew a lot of quilts for my grandchildren. Do you have any kids?\newline \textbf{human: }Aww that's sweet! Yes, I have one son. \newline \textbf{bot: }How old is your son? I have two daughters. Your daughters are a handful!\newline \textbf{human: }He's four and yes they are a handful! How old are your daughters?\newline \textbf{bot: }My daughters are 4, 5 and 8. I love my daughters to death, but they're exhausting!                                                   & Numerical Reasoning \\ \midrule
\textbf{human: }Hi!\newline \textbf{bot: }Hi there! How are you? What do you do for a living? I'm a college student.\newline \textbf{human: }I'm good, thankyou! I'm an electrician. What do you study?\newline \textbf{bot: }It is in education. I just moved to an apartment near my job. What are you studying?                                                                                                                                                                                                                                                                                                   & \textbf{human: }Hi!\newline \textbf{bot: }Hi there! Are you okay? What do you do for a living? I'm a college student.\newline \textbf{human: }I'm good, thankyou! I'm an electrician. What do you study?\newline \textbf{bot: }I study education. I just moved to an apartment near my job. What are you studying?                                                                                                                                                                                                                                                                                                                  & Logical Reasoning    \\ \midrule
\textbf{human: }I've never left the US, I think I'm afraid to travel internationally though.\newline \textbf{bot: }I can understand that. I have never been to the united states. I would love to though.\newline \textbf{human: }Oh I thought you lived in the South United States.  Where do you live?\newline \textbf{bot: }I live on the south side of the state. It is very cold in the winter, though.                                                                                                                                                                                                         & \textbf{human: }I've never left the US, I think I'm afraid to travel internationally though.\newline \textbf{bot: }I can understand being afraid to travel internationally. I have never been to the united states. I would love to travel internationally though.\newline \textbf{human: }Oh I thought you lived in the South United States.  Where do you live?\newline \textbf{bot: }I live on the south side of the state. It is very cold in the winter though.                                                                                                                                                                & Common Sense Reasoning         \\ \midrule
\textbf{human: }What did you do today? I have an adult son.\newline \textbf{bot: }I went to the mall with my mom. She's my best friend. How old is your son?\newline \textbf{human: }My son is 29. How many children do you have?\newline \textbf{bot: }I have three kids. They're all grown up now, but they're still my best friends.\newline \textbf{human: }Do your kids live with you?\newline \textbf{bot: }Yes, they live with me and my husband. What do you like to do in your spare time?\newline \textbf{human: }Have you been shopping lately? I love shopping.\newline \textbf{bot: }No, I don't have much time for shopping. I spend most of my time with my kids. & \textbf{human: }What did you do today? I have an adult son.\newline \textbf{bot: }I went to the mall with my mom. My mom is my best friend. How old is your son?\newline \textbf{human: }My son is 29. How many children do you have?\newline \textbf{bot: }I have three kids. My kids are all grown up now, but my kids are still my best friends.\newline \textbf{human: }Do your kids live with you?\newline \textbf{bot: }Yes, my kids live with me and my husband. What do you like to do in your spare time?\newline \textbf{human: }Have you been shopping lately? I love shopping.\newline \textbf{bot: }No, I don't have much time for shopping. I spend most of my time with my kids. & Hard to Judge        \\ \bottomrule
\end{tabular}}
\caption{Examples that are wrongly predicted as negatives before and after automatic rewriting. }
\label{tab:appendix-still-wrong-examples}
\end{table*}

\begin{table*}[t]
\centering
\small
\resizebox{0.99\textwidth}{!}{
\begin{tabular}{p{7cm}|p{7cm}}
\toprule
\textbf{Original Dialogue}                                                                                                                                                                                                                                                                                                                                                                                                                                                                                    & \textbf{Rewritten Dialogue}                                                                                                                                                                                                                                                                                                                                                                                                                                                                                                                                                                                                 \\ \midrule
\textbf{human:} Hi!\newline \textbf{bot:} Hey!\newline \textbf{human:} How are you today?\newline \textbf{bot:} Great! And you?\newline \textbf{human:} I am doing great as well! What do you do for a living?\newline \textbf{bot:} I injured in an accident.\newline \textbf{human:} When and what happened?\newline \textbf{bot:} I had a sports injury.\newline \textbf{human:} What kind of accident?\newline \textbf{bot:} Broke an ankle. Very painful.\newline \textbf{human:} What sport were you in?\newline \textbf{bot:} I compete in rowing.\newline \textbf{human:} How did you break an ankle from rowing?\newline \textbf{bot:} A four wheeler accident. & \textbf{human:} Hi!\newline \textbf{bot:} Hey!\newline \textbf{human:} How are you today?\newline \textbf{bot:} I am great today! And how are you?\newline \textbf{human:} I am doing great as well! What do you do for a living?\newline \textbf{bot:} I was injured in an accident.\newline \textbf{human:} When and what happened?\newline \textbf{bot:} I had a sports injury that caused me to be injured.\newline \textbf{human:} What kind of accident?\newline \textbf{bot:} I broke an ankle. My injury was very painful.\newline \textbf{human:} What sport were you in?\newline \textbf{bot:} I compete in rowing when I broke my ankle.\newline \textbf{human:} How did you break an ankle from rowing?\newline \textbf{bot:} I broke my ankle in a four wheeler accident. \\ \midrule
\textbf{human:} Hi!\newline \textbf{bot:} Hello, how are you? I'm a huge fan of star wars. How about you?\newline \textbf{human:} I have never seen a Star Wars movie.\newline \textbf{bot:} Neither have I. What do you like to do in your spare time? I like coffees.                                                                                                                                                                                                                                            & \textbf{human:} Hi!\newline \textbf{bot:} Hello, how are you? I'm a huge fan of star wars. How about you?\newline \textbf{human:} I have never seen a Star Wars movie.\newline \textbf{bot:} I have never seen a Star Wars movie. What do you like to do in your spare time? I like coffees.                                                                                                                                                                                                                                                                                                                                     \\ \midrule
\textbf{human:} Hi!\newline \textbf{bot:} What do you like to do in your spare time? I like to run and play soccer.\newline \textbf{human:} I like to work on cars in my spare time. Do you play any sports?\newline \textbf{bot:} No, I don't. I'm not very athletic. What kind of cars do you work on?                                                                                                                                                                                                           & \textbf{human:} Hi!\newline \textbf{bot:} What do you like to do in your spare time? I like to run and play soccer in my spare time.\newline \textbf{human:} I like to work on cars in my spare time. Do you play any sports?\newline \textbf{bot:} No, I don't play any sports. I'm not very athletic. What kind of cars do you work on?                                                                                                               \\ \bottomrule                                                                                                                                                                       
\end{tabular}}
\caption{Samples that are wrongly predicted as negatives before rewriting but get corrected after rewriting.}
\label{tab:appendix-get-corrected-samples}
\end{table*}




\end{document}